\documentclass[11pt, preprint]{article}

\usepackage{acl}
\usepackage{times}
\usepackage{latexsym}
\usepackage[T1]{fontenc}
\usepackage[utf8]{inputenc}
\usepackage{microtype}
\usepackage{inconsolata}
\usepackage{graphicx}

\usepackage{acronym}
\usepackage{svg}
\usepackage{amsmath}
\usepackage{rotating}
\usepackage{multirow}
\usepackage{placeins}
\usepackage[normalem]{ulem}
\useunder{\uline}{\ul}{}

\title{Does task decomposition improve automatic NLG evaluation?}

\author{
    Sebastian Steindl$^1$ \quad Nikos Voskarides$^1$ \quad Alberto Gasparin$^2$ \quad Diego Marcheggiani$^1$ \\
    \\
    $^1$Amazon, Barcelona, Spain \quad $^2$Amazon, Berlin, Germany \\
    \texttt{\{sebstei,nvvoskar,marchegg\}@amazon.es}, \texttt{albgas@amazon.de}
  }

\begin{document}
\acrodef{NLG}{Natural Language Generation}
\acrodef{NLP}{Natural Language Processing}
\acrodef{AutoEval}{Automatic Evaluation}
\acrodef{LLM}{Large Language Model}
\acrodef{IAA}{Inter-Annotator Agreement}
\acrodef{ICL}{In-Context Learning}
\acrodef{LLMaJ}{LLM-as-a-judge}

\maketitle

\begin{abstract}
The LLM-as-a-judge (LLMaJ) framework has emerged as a promising solution for cheap, reproducible, reference-free Natural Language Generation (NLG) evaluation.
Prior work seeks to improve LLMaJ by decomposing evaluation tasks into simpler sub-tasks.
In this work, we systematically compare LLMaJ methods with and without decomposition on multiple NLG datasets.
We find no evidence that LLMaJ with task decomposition leads to performance gains over a fair baseline that does not use decomposition.
Instead, we find that previously reported performance gains in decomposition-based LLMaJ stem from using human labels as training data, and not task decomposition itself.
Also, we find that, when human labels are available, LLMaJ without using task decomposition can perform comparably to human annotators.
\end{abstract}

\section{Introduction}

Evaluating the output quality of automatic \ac{NLG} methods is intricate: widely used reference-based metrics such as ROUGE~\cite{lin-2004-rouge} and BERTScore~\cite{BERTScore} rely on expensive human annotations, operate on surface-level features, and correlate poorly with human judgments~\cite{gehrmann2023repairing}.
The \ac{LLMaJ} framework~\cite{LLMJudge,wang-etal-2023-chatgpt} has thus gained attention as a reference-free alternative~\cite{LLM_judge_NLG_survey}. A recent line of work proposes decomposing each evaluation criterion into simpler sub-criteria, aiming to improve alignment with human annotations and reduce variance across models and executions~\cite{min-etal-2023-factscore,saha-etal-2024-branch,liu-etal-2024-hd,lee-etal-2025-checkeval}.

\begin{figure*}[tbh]
    \centering
    \includegraphics[scale=0.55]{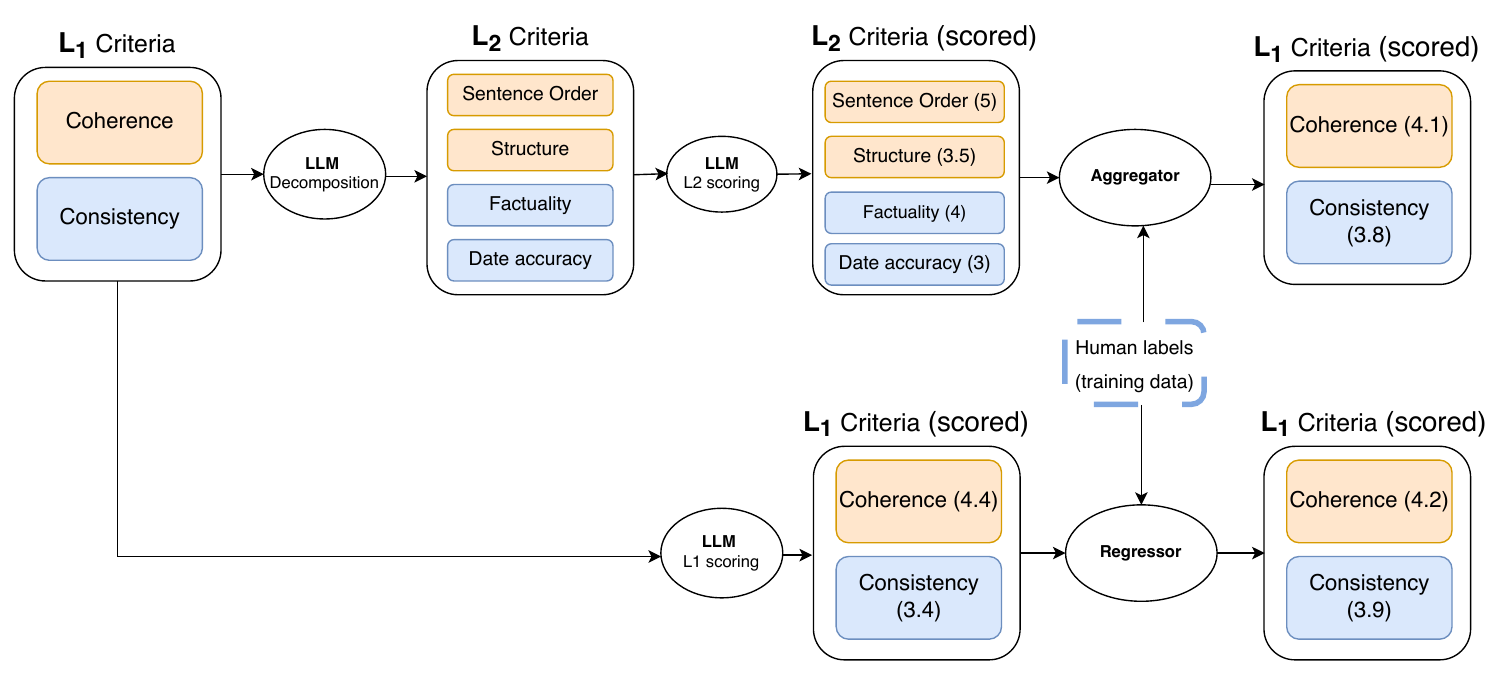}
    \caption{Visualization of decomposition-based \ac{LLMaJ} (upper part), and Direct Prediction \ac{LLMaJ} (lower part).
    Human labels are optional: they are used by HD-Eval, can be used by Direct Prediction but are not used by CheckEval. 
    The input text $t$ is not shown.}
    \label{fig:mainfig}
\end{figure*}

In this paper, we perform a systematic analysis of decomposition-based \ac{LLMaJ} (Figure~\ref{fig:mainfig}).
These methods follow a three-stage pipeline: an LLM decomposes a given evaluation criterion (e.g., \texttt{Coherence}) into subcriteria, scores each subcriterion, and an aggregator produces a single score, either by using a learned regressor~\cite{liu-etal-2024-hd} or by a simple average of subcriteria scores~\cite{lee-etal-2025-checkeval}.
We ask whether this decomposition step is actually beneficial. To answer this, we construct a strong baseline that directly predicts the score for a given criterion without decomposition, granting it access to the same human labels used by \citet{liu-etal-2024-hd} to ensure a fair comparison (we also report results without human labels).
Additionally, we attempt to improve decomposition-based \ac{LLMaJ} itself through more principled decomposition logic and \ac{ICL}~\cite{brown2020language}.

Our experimental results show no evidence of decomposition-based \ac{LLMaJ} being superior to the  baseline that does not use decomposition.
In fact, our analysis showed that previously reported gains of decomposition-based \ac{LLMaJ} are not due to task simplification by decomposition, but rather due to using human labels as training data.
In summary, our main contributions are: i) a systematic comparison of decomposition-based and direct prediction \ac{LLMaJ} approaches, ii) insights into why decomposition-based \ac{LLMaJ} is not beneficial, iii) evidence that direct prediction \ac{LLMaJ} can reach human-level performance on some evaluation criteria when using human ratings as training data.

\section{Problem Statement}
Given a text $t$ to be evaluated according to an evaluation criterion $d$ (e.g. \texttt{Coherence}), we define the annotation score from an \ac{LLMaJ} $f$ as $s = f(t, d)$.
The goal is to align $s$ as closely as possible to the human score $s^* = human(t, d)$.
We follow the notation from \citet{liu-etal-2024-hd} and denote with $L_1$ the set of evaluation criteria rated by humans and with $L_n$ the set of sub-criteria after $n-1$ decomposition steps. For example, the $L_1$ criterion \texttt{Coherence} could be broken down into the $L_2$ sub-criteria \texttt{Sentence Order}, \texttt{Structure}, and others.

We study two approaches for predicting a score $s$ for an $L_1$ criterion: decomposition-based, where $L_1$ is decomposed to sub-criteria and then aggregated back to $L_1$ to make a prediction  (Section \ref{sec:decomp}), and direct prediction, where $L_1$ is predicted directly without decomposition (Section \ref{sec:direct-prediction}). 

\section{Decomposition-based LLMaJ}
\label{sec:decomp}

Fig. \ref{fig:mainfig} (upper part) shows an overview of the decomposition-based \ac{LLMaJ}.
Given a text $t$, it first uses an LLM to decompose an $L_1$ evaluation criterion into $L_2$ sub-criteria and then score each sub-criterion.
Then, it aggregates the scores of sub-criteria to output a single score $s$.
We study decomposition-based approaches both when human labels are used to train an aggregator, and when they are withheld.  
Note that human annotations are only available at the $L_1$ level; no ground-truth exists for the decomposed sub-criteria.

\subsection{Existing approaches}
Existing approaches differ along three axes: the \emph{structure} of the decomposition (flat vs.\ hierarchical), the \emph{scale} used for scoring (binary vs.\ ordinal), and the \emph{aggregation} mechanism
(label-free vs.\ learned).

We focus on HD-Eval~\cite{liu-etal-2024-hd} and CheckEval~\cite{lee-etal-2025-checkeval} as representative instantiations of this template, since they decompose into a general set of sub-criteria rather than generating instance-level checklists~\cite{wei2025rocketeval}. Both are evaluated on the same well-known NLG benchmarks, enabling direct comparison.
Related but out of scope is LLM-Rubric~\cite{hashemi-etal-2024-llm}, which defines evaluation criteria but trains a calibrator to map LLM scores to human scores without a decomposition step, making it a calibration method rather than a decomposition method. Also, \citet{li2024decompose} focus on pairwise evaluation whereas we consider a single instance at a time.
\paragraph{HD-Eval~\cite{liu-etal-2024-hd}}
HD-Eval decomposes an $L_1$ evaluation criterion into $L_2$ and $L_3$ sub-criteria in a hierarchical manner, which are then scored by the LLM on an ordinal scale (e.g., 1-5).
As an aggregator, HD-Eval uses a regressor trained on human labeled data (that  are assumed to be available at the $L_1$ level) to output the score $s$.\footnote{We do not consider \citet{liu-etal-2024-calibrating} since it has consistently inferior performance than HD-Eval.}
In our reproduction of HD-Eval we use the $L_2$ and $L_3$ sub-criteria as input to the regressor.\footnote{
Note that, unless stated otherwise, we do not use direct $L_1$ predictions as input as they do in the original paper, since that would not allow us to disentangle the effect of decomposition and direct prediction. Nevertheless, we find that also using $L_1$ criteria as input does not lead to performance gains.}

\paragraph{CheckEval~\cite{lee-etal-2025-checkeval}}
CheckEval decomposes an $L_1$ evaluation criterion into $L_2$ sub-criteria that are binary yes/no questions.
Differently from HD-Eval, they do not use human labeled data to aggregate the $L_2$ scores.
Instead, the aggregator simply outputs the proportion of positively answered binary questions as the score $s$.
This approach is fully equivalent to TICK \cite{cook2024ticking}.

\subsection{Proposed Extensions}
\label{sec:extensions}
We investigate two extensions to HD-Eval~\cite{liu-etal-2024-hd}, in an attempt to further improve the performance of decomposition-based \ac{LLMaJ}.
First, we introduce Atomic, Observable, and Independent (AOI) decomposition, which creates sub-criteria that are meant to be uncorrelated (Atomic, Independent) and clearly derivable from the text (Observable) (see Fig. \ref{fig:prompt_decompAOI}).
Second, we apply \ac{ICL} during decomposition \cite{brown2020language,sanh2022multitask,dong-etal-2024-survey} to guide the LLM to infer human-aligned sub-criteria.\footnote{Implementation details in Appendix \ref{sec:appendix_extraDecomp}.}
Both can be seen as additional decomposition variants that extend the reasoning underlying the decomposition approach. Thus, one would hypothesize that they lead to better decompositions and consequently better alignment to humans.
\section{Direct Prediction LLMaJ}
\label{sec:direct-prediction}

We establish a strong, Direct Prediction \ac{LLMaJ} baseline that predicts a score $s$ for an $L_1$ evaluation criterion without decomposition (Fig. \ref{fig:mainfig}, lower part).
Following common practices~\cite{LLM_judge_NLG_survey}, the prompt contains the criterion name and definition, the rating scale, a CoT-inducing statement (Fig.~\ref{fig:prompt_direct}), and five \ac{ICL}~\cite{brown2020language} examples showcasing human evaluations at low, medium, and high scores.\footnote{Here, ICL is used for directly predicting $L_1$ scores, not for guiding decomposition as in Section~\ref{sec:extensions}.}
Since this baseline directly predicts the target $L_1$ score, no aggregation step is needed.
However, for fair comparison with HD-Eval, we also train a regressor on the uni-dimensional LLM output using human labels, which shifts predictions closer to the human distribution without influencing LLM inference.\footnote{For fair comparison with CheckEval, which uses no human labels, we also report results without the regressor.}

\section{Experimental Setup}

\paragraph{Datasets}
Following \citet{liu-etal-2024-hd} and \citet{lee-etal-2025-checkeval}, we use SummEval \cite{fabbri-etal-2021-summeval} and TopicalChat \cite{gopalakrishnan2019topical} as the main datasets. 
Additionally, we use Seahorse \cite{clark-etal-2023-seahorse} as an alternative summarization dataset with multiple evaluation criteria.
\paragraph{Metrics} To evaluate the quality of \ac{LLMaJ} methods, we measure their alignment to human labels for $L_1$ criteria.
We follow standard practices and report alignment as measured by correlation metrics.
Concretely, we report Spearman's $\rho$.\footnote{Pearson's $r$ and Kendall's $\tau$ shown in the full results in the Appendix.}
Moreover, we adopt the alt-test \cite{calderon-etal-2025-alternative} for SummEval and TopicalChat, where we have access to scores from multiple (three) annotators.
We report the Average Advantage Probability (AP) and Winning-Rate (WR), which are both in $[0, 1]$ (higher is better).
AP represents the probability that the LLM annotations are as good as or better than those of a randomly chosen annotator, and can be used to compare judges against each other.
WR is the percentage of annotators against which the \ac{LLMaJ} ``wins''. 
For the Seahorse dataset it is not possible to use the alt-test since only one annotation per sample is available.
We thus report accuracy and Krippendorf's $\alpha$, and calculate \ac{IAA} to compare to human labels, following \citet{clark-etal-2023-seahorse}.

\paragraph{Implementation Details} We use Claude-4 (Sonnet)~\cite{claude4} with temperature $t=0$ as the base LLM for our main results. We also provide results with Qwen3-32B \cite{yang2025qwen3} and GPT-OSS-120B~\cite{agarwal2025gpt} in the Appendix \ref{sec:appendix_otherModels}. We observe generally the same trends across these models.
To ensure the best possible comparison, we re-implement HD-Eval~\cite{liu-etal-2024-hd} and CheckEval~\cite{lee-etal-2025-checkeval} with Claude-4 using their provided code or prompts, and additionally also include the results reported in the original publications.
We use a 50/50 train/test split as in~\cite{liu-etal-2024-hd}.
We follow HD-Eval \cite{liu-etal-2024-hd} and experiment with different regressors: Linear Regression, Decision Tree, Random Forest, Multilayer Perceptron (details in Appendix~\ref{sec:aggregator-implementation-details}). We report results from the best regressor per setting. The best regressor is selected for every evaluated setting individually and choesn by the combination of Spearman's $\rho$ and AP score. The same regressor is used for all evaluation dimensions to ensure fair comparison. The best regressors are shown in the full results in Tables \ref{tab:result_perdim_summeval} and \ref{tab:result_perdim_topical}.
All methods under comparison output floats.\footnote{
Float outputs gain a numerical advantage over integer outputs through tie elimination and better fit to float ground truths; we provide results both with and without rounding to ensure fair comparison (Appendix \ref{sec:appendix_floatsTiesRounding}).}
We report results on a single run, but did not observe significant variations across runs.

\begin{table*}[t]
\centering
\resizebox{0.95\textwidth}{!}{%
\begin{tabular}{lcc|ccc|ccc}
\hline
\multirow{2}{*}{\bf Method} & \multirow{2}{*}{\bf Decomp.} & \multirow{2}{*}{\bf Human labels}  & \multicolumn{3}{c|}{\bf SummEval} & \multicolumn{3}{c}{\bf TopicalChat} \\
 &  &  & $\rho \uparrow$ & AP $\uparrow$ & WR $\uparrow$ & $\rho\uparrow$ & AP $\uparrow$ & WR $\uparrow$ \\ \hline
HD-Eval$^\ast$ & yes & yes & 0.535 & – & – & 0.638 & – & – \\
HD-Eval (Claude 4) & yes & yes & 0.567 & 0.833 & \textbf{1.0} & 0.563 & 0.848 & 0.58 \\
HD-Eval (Claude 4) + $L_1$ & yes & yes & 0.560 & 0.828 & \textbf{1.0} & 0.582 & 0.843 & 0.58 \\
CheckEval$^\ast$ (Mistral-Large) & yes & no & 0.549 & – & – & 0.645 & – & – \\
CheckEval$^\ast$ (GPT-4o) & yes & no & 0.504 & – & – & 0.640 & – & – \\
CheckEval (Claude 4) & yes & no & 0.411 & – & – & 0.428 & – & – \\ \hline
ICL Decomposition & yes & yes & 0.567 & 0.842 & \textbf{1.0} & 0.575 & 0.840 & \textbf{1.0} \\
AOI Decomposition & yes & yes & \textbf{0.586} & 0.831 & \textbf{1.0} & 0.574 & 0.839 & \textbf{1.0} \\ \hline
Direct Prediction & no & no & 0.553 & 0.593 & 0.5 & 0.701 & 0.864 & 0.5 \\
Direct Prediction (+ICL) & no & no & 0.552 & 0.638 & 0.75 & \textbf{0.716} & 0.870 & 0.58 \\
Direct Prediction & no & yes & 0.560 & 0.835 & \textbf{1.0} & 0.672 & 0.885 & \textbf{1.0} \\ 
Direct Prediction (+ICL) & no & yes & 0.545 & \textbf{0.846} & \textbf{1.0} & 0.687 & \textbf{0.888} & \textbf{1.0} \\ \hline
\end{tabular}
} 
\caption[Main results of our study.]{Main results of our study. We show the results averaged across all criteria. We report sample-wise Spearman's $\rho$,  Advantage Probability (AP), and Win-Rate (WR). Pearson's $r$ and Kendall's $\tau$ are shown with the per-criterion results in the Appendix \ref{sec:appendix_perCritResult}. 
Column "Decomp." refers to whether decomposition is performed.
Column "Human Labels" refers to whether human labels are used either during the aggregation step (HD-Eval) or for aligning direct prediction scores.
Note that we do not report AP and WR for CheckEval since the output is on a different scale than the human responses, making their calculation not meaningful.
$^\ast:$As originally reported.
}
\label{tab:main_table}
\end{table*}

\section{Results and Discussion}
\subsection{Decomposition VS Direct Prediction}

Table \ref{tab:main_table} shows the performance of different \ac{LLMaJ} methods on the SummEval and TopicalChat datasets. 
First, we see that Direct Prediction combinations perform better or only slightly worse than the best decomposition-based approaches, with one of them achieving the best overall AP on both datasets.
Also, we see that even the Direct Prediction combination that does not use human labels is on par or better than CheckEval.\footnote{Note that our re-production of CheckEval with the original code but with Claude-4 led to worse results than originally reported. Nevertheless, the conclusions we make here hold even with their best reported results.}
Furthermore, our proposed extensions to decomposition-based approaches (ICL and AOI) show some marginal improvements but do not consistently outperform Direct Prediction.
Our second main finding is that using human labels for Direct Prediction leads to improved performance. Part of the improvement can be attributed to numerical benefits from the regressor outputting floating-point numbers. More details in Appendix \ref{sec:appendix_floatsTiesRounding}.
Results on Seahorse (Table \ref{tab:seahorse} in Appendix \ref{sec:appendix_resultSeahorse}) confirm the trend: none of the decomposition-based approaches evaluated in this work consistently outperforms the Direct Prediction baseline.
To further test whether decomposition from $L_1$ criteria improves performance through task simplification, we modify HD-Eval to be $L_1$-agnostic: we ask the LLM to generate 25 general quality criteria for the task without specifying the target $L_1$ criterion, then train the regressor on human labels as usual.
This $L_1$-agnostic variant performs on par with both standard decomposition and Direct Prediction (Table~\ref{tab:L0_decomp}, Appendix~\ref{sec:appendixL0decomp}), indicating that HD-Eval's gains stem from the learned aggregation rather than decomposition itself.
This might suggest that the LLM already possesses an internal understanding of quality that the regressor can align to specific $L_1$ criteria.

\subsection{How far is Direct Prediction LLMaJ from human-level performance}
Table \ref{tab:main_table} shows that, even without using human labels, Direct Prediction achieves a $\text{WR} \geq 0.5$ on both datasets, which is the threshold at which LLMs are considered as good as human annotators~\cite{calderon-etal-2025-alternative}.
However, when we consider per-criterion scores instead of the average across all criteria (Tables~\ref{tab:result_perdim_summeval} and \ref{tab:result_perdim_topical} in Appendix~\ref{sec:appendix_perCritResult}), we identify criteria where Direct Prediction \ac{LLMaJ} outperforms human annotators even without using human labels (e.g. \texttt{Coherence} in SummEval), whereas for other dimensions it does not (e.g. \texttt{Relevance}). 
When using human labels, Direct Prediction can reach a WR of 100\% on both datasets, which means that the LLM is closer to the average human rating than the humans individually~\cite{calderon-etal-2025-alternative}.
Results on Seahorse are similar (see Table \ref{tab:seahorse} in Appendix \ref{sec:appendix_resultSeahorse}). We see that, while on average the LLMs still lack behind humans, the difference is mostly due to two criteria, namely \texttt{Grammar} and \texttt{Main Ideas}. On the other three criteria, Direct Prediction is close enough to humans to consider it a valid alternative.\footnote{A more detailed analysis is provided in Appendix \ref{sec:appendix_resultSeahorse}.}

\section{Conclusion}
We studied two decomposition-based \ac{LLMaJ} methods for NLG evaluation across three datasets and multiple LLMs, finding that they do not consistently outperform a fair Direct Prediction baseline we designed.
Our analysis shows that HD-Eval's gains stem from access to human labels, not from task decomposition as previously claimed.
Thus, despite the intuitive appeal of decomposition~\cite{burchardt-2013-multidimensional}, we find no compelling evidence that it improves \ac{LLMaJ}.
Finally, we show that Direct Prediction with human labels can reach human-level performance on some evaluation tasks on both SummEval and TopicalChat.
Future work might study how decomposition-based evaluation performs in settings where humans rate not only the original criteria, but also the decomposed criteria.  
\section*{Limitations}
First, the number of NLG tasks and datasets that we consider is limited, but is consistent with the prior work we base our study upon and diverse enough to support our conclusions.
While we find no compelling evidence that decomposition is beneficial for typical NLG tasks, this finding does not necessarily extend to complicated, multi-step reasoning tasks.
Further, studying \ac{LLMaJ} from the perspective of bias, fine-tuning, and human-LLM-collaboration \cite{LLM_judge_NLG_survey} is out of the scope for our work.
Moreover, our replication of the CheckEval results using their published code but replacing the LLM with Claude-4 to be consistent with the rest of the methods led to much worse performance than reported by \citet{lee-etal-2025-checkeval}. We want to stress that our findings remain true, even when comparing to the better, originally reported performance.
Lastly, a general limitation of our study is that we cannot determine whether the evaluated LLMs were exposed to our benchmark datasets during pre-training.

\bibliography{custom}

\FloatBarrier

\appendix

\section{Results on Seahorse}\label{sec:appendix_resultSeahorse}
For the Seahorse dataset, we do not have multiple annotations for the examples, and so there is no principled way of considering the cost-benefit tradeoff as in the alt-test. Therefore, we can only rely on the \ac{IAA} published by \citet{clark-etal-2023-seahorse} as a comparison. However, we see again that the decompositions do not improve above the baseline. In most cases, they are actually worse. This might be partly because the criteria are already relatively atomic. 
The metrics used are the accuracy and Krippendorf's $\alpha$. While the accuracy has limited meaningfulness, since i) the criteria are binary, increasing per-chance agreement, and ii) some criteria show a significant class-imbalance. Therefore, Krippendorf's $\alpha$ is the main metric. We still report accuracy to enable comparison to the results from \citet{clark-etal-2023-seahorse}.
While on average, the LLMs still lack behind the humans in the Krippendorf's $\alpha$, the difference comes mostly from two criteria, namely \textit{Grammar} and \textit{Main Ideas}. On the other three criteria, the LLM is at minimum close enough to humans to consider it a valid alternative.  
Tab. \ref{tab:seahorse} shows the results on the Seahorse dataset.

\begin{table*}
\centering
\resizebox{0.95\textwidth}{!}{%
\begin{tabular}{lcccccccccc|cc}
\hline
Method & \multicolumn{2}{c}{Grammar} & \multicolumn{2}{c}{Attributable} & \multicolumn{2}{c}{Main Ideas} & \multicolumn{2}{c}{Conciseness} & \multicolumn{2}{c}{Repetition} & \multicolumn{2}{c}{Average} \\
 & Acc. & $\alpha$ & Acc. & $\alpha$ & Acc. & $\alpha$ & Acc. & $\alpha$ & Acc. & $\alpha$ & Acc. & $\alpha$ \\ \hline
Human IAA & 0.94 & 0.87 & 0.95 & 0.35 & 0.69 & 0.47 & 0.61 & 0.40 & 0.69 & 0.41 & 0.78 & 0.50 \\ \hline
Direct Prediction & \textbf{0.81} & \textbf{0.33} & \textbf{0.74} & \textbf{0.48} & {\ul 0.65} & 0.27 & \textbf{0.72} & {\ul 0.39} & \textbf{0.95} & \textbf{0.68} & \textbf{0.77} & \textbf{0.43} \\
Decomposition (AOI) & 0.76 & 0.21 & 0.66 & 0.38 & {\ul 0.65} & 0.28 & \textbf{0.72} & \textbf{0.40} & {\ul 0.93} & {\ul 0.62} & 0.74 & 0.38 \\
Decomposition (ICL) & {\ul 0.78} & {\ul 0.27} & {\ul 0.72} & {\ul 0.45} & \textbf{0.66} & 0.30 & 0.71 & 0.38 & 0.91 & 0.54 & {\ul 0.75} & {\ul 0.39} \\ \hline
\end{tabular}
} 
\caption{Results on the Seahorse \cite{clark-etal-2023-seahorse} dataset. Reporting Accuracy (Acc.) and Krippendorf's $\alpha$. All results use vertical aggregation.}
\label{tab:seahorse}
\end{table*}

\section{Further Qualitative Examples}
\begin{figure*}[!htb]
    \centering
    \includegraphics[width=0.5\textwidth]{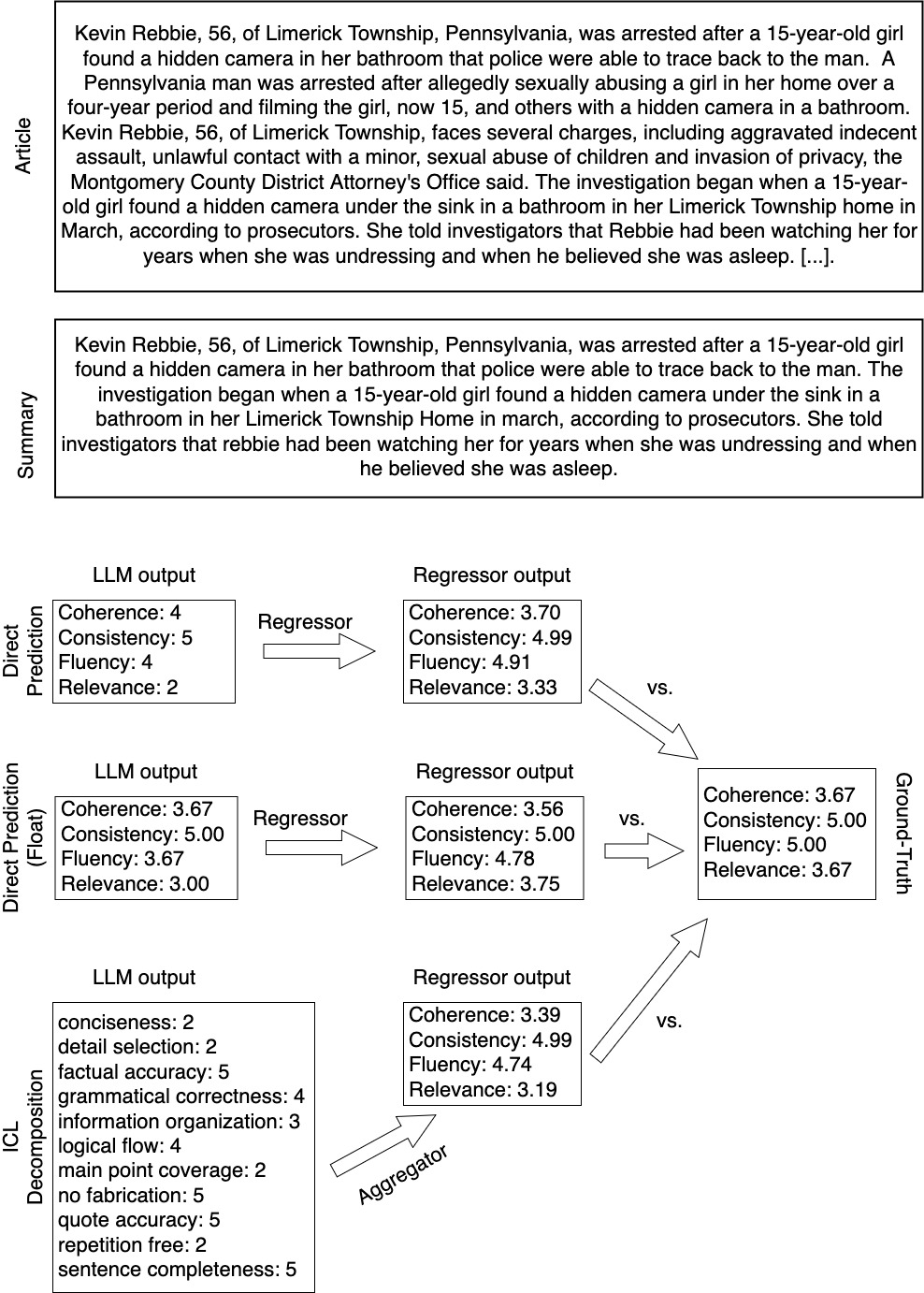}
    \caption{Further example from the SummEval dataset, showcasing the outputs from a direct integer prediction (top), direct floating point prediction (middle) and the ICL-decomposition (bottom). After regression / aggregation step, the outputs are better aligned both with each other, and with the ground-truth.}
    \label{fig:secondExample}
\end{figure*}

In Fig. \ref{fig:secondExample} we show another qualitative example that shows outputs from different methods for a given input before and after applying the Regressor or the Aggregator. 

\section{L1-agnostic Decomposition}
\label{sec:appendixL0decomp}
Table~\ref{tab:L0_decomp} shows the results for $L_1$-agnostic decomposition, i.e., asking the LLM to come up 25 criteria or questions that can be used to evaluate the quality of a text, ignoring the $L_1$ criteria.

\begin{table*}[tbh]
\centering
\begin{tabular}{ll|ccc|ccc}
\hline
 &  & \multicolumn{3}{c|}{SummEval} & \multicolumn{3}{c}{TopicalChat} \\
Method & ICL & $\rho$ & AP & WR & $\rho$ & AP & WR \\ \hline
ICL Decomposition & no & {\ul 0.594} & {\ul 0.853} & \textbf{1.0} & 0.604 & 0.852 & \textbf{1.0} \\
AOI Decomposition & no & \textbf{0.608} & 0.850 & \textbf{1.0} & {\ul 0.622} & {\ul 0.862} & 0.75 \\
$L_1$-agnostic Decomposition & no & 0.579 & \textbf{0.857} & \textbf{1.0} & 0.621 & 0.856 & \textbf{1.0} \\
Direct Prediction & yes & 0.566 & 0.842 & \textbf{1.0} & \textbf{0.704} & \textbf{0.90} & \textbf{1.0} \\ \hline
\end{tabular}
\caption{Comparison of $L_1$-agnostic Decomposition to ICL and AOI decomposition, and direct prediction. All aggregations performed horizontally. All methods use human labels.}
\label{tab:L0_decomp}
\end{table*}

\section{Implementation Details} \label{sec:appendix_implDetails}

\subsection{ICL and AOI decomposition} \label{sec:appendix_extraDecomp}
We investigate two modifications to the HD-Eval framework, in an attempt to further improve the performance of decomposition approaches.
Both modifications are motivated by the intuition of the frameworks working mechanism.
First, we use \ac{ICL} \cite{brown2020language,sanh2022multitask,dong-etal-2024-survey} while decomposing (ICL Decomposition). The rationale is that seeing not only the description of the evaluation criteria, but also examples of how concrete examples were rated, allows the LLM to infer what features the annotators paid attention to. Thus, the fine-grained criteria can be better aligned to the criteria the raters used to find their score.

As a second approach, we propose a novel decomposition, which we call atomic, observable and independent (AOI).
It uses three prompts in sequence to create sub-criteria that are AOI, aiming to maximize their usefulness for the aggregation step (cf. Fig. \ref{fig:prompt_decompAOI} in the Appendix).
The AOI-decomposition additionally tries to enforce sub-criteria that are simpler to evaluate, more objective, and more informative in the aggregation step. First, the LLM is tasked to decompose the evaluation criteria into atomic sub-criteria, that is, criteria that measure only one thing. This should simplify the evaluation from the perspective of the \ac{LLMaJ}.
Next, it is prompted to make them directly observable from the text. This is aimed to increase the objectivity.
Finally, it is prompted to make the now atomic and observable sub-criteria as independent from each other as possible. The motivation for this is to increase the information contained in those sub-criteria when used as input to the regression model, by reducing the overlap (correlation) between them.
While the decomposition step could be applied recursively, the ablations showed that the gain after $L_2$ becomes marginal \cite{liu-etal-2024-hd}. 
For the ICL and AOI decomposition, we remain at $L_2$. When reproducing the HD-Eval scores with Claude-4, we do not provide the direct $L_1$ predictions to the aggregator.
Contrary to HD-Eval, we do not aggregate the values horizontally, but vertically.  That is, only the children of a given $L_1$ criterion are used to train the regressor, instead of all decomposed sub-criteria. This practice is more aligned with the theoretical motivation of task-decomposition. A more detailed discussion and performance comparison of both approaches is given in the Appendix \ref{sec:appendix_horizontalVertical}.

For the alt-test, we follow the instructions from \citet{calderon-etal-2025-alternative} and set the False Discovery Rate $q := 0.05$ and the cost-benefit penalty to $\epsilon := 0.2$ and $\epsilon := 0.1$ for SummEval and TopicalChat, respectively. The difference in $\epsilon$ represents the fact that in SummEval we compare to expert annotators and in TopicalChat to crowd-workers.

\subsection{Vertical or horizontal aggregation} \label{sec:appendix_horizontalVertical}

\begin{table*}[tbh]
\centering
\resizebox{0.95\textwidth}{!}{%
\begin{tabular}{lllll|ccc|ccc}
\hline
 & Aggregation & Human &  & Outputs & \multicolumn{3}{c|}{SummEval} & \multicolumn{3}{c}{TopicalChat} \\
Method & type & labels & ICL & floats & $\rho$ & AP & WR & $\rho$ & AP & WR \\ \hline
\multirow{2}{*}{ICL Decomposition} & vertical & yes & no & yes & 0.567 & 0.842 & \textbf{1.0} & 0.575 & 0.840 & \textbf{1.0} \\
 & horizontal & yes & no & yes & 0.594 & \textbf{0.853} & \textbf{1.0} & 0.604 & 0.852 & 0.75 \\
\multirow{2}{*}{AOI Decomposition} & vertical & yes & no & yes & 0.586 & 0.831 & \textbf{1.0} & 0.574 & 0.839 & 0.33 \\
 & horizontal & yes & no & yes & \textbf{0.608} & 0.850 & \textbf{1.0} & 0.622 & 0.862 & 0.75 \\ \hline
\multirow{2}{*}{Direct Float Prediction} & vertical & yes & yes & yes & 0.545 & \textbf{0.846} & \textbf{1.0} & 0.687 & 0.888 & \textbf{1.0} \\
 & horizontal & yes & no & yes & 0.566 & 0.842 & \textbf{1.0} & \textbf{0.704} & \textbf{0.896} & \textbf{1.0} \\ \hline
\end{tabular}
} 
\caption{Vertical vs. horizontal aggregation.}
\label{tab:verticalHorizontal}
\end{table*}

HD-Eval performs the aggregation in what we call horizontal fashion. That is, while training one regressor for each of the $L_1$ criteria, they use \textit{all} decomposed sub-criteria as input. Thus, the regressor uses information from e.g. the prediction of \textit{Grammar}, which was decoded from \textit{Fluency}, to predict the score of \textit{Consistency}. 
In our understanding, this practice is not aligned with the idea of decomposition. 
Instead, to follow the principles of breaking down a task into sub-tasks, whose can be aggregated into a solution to the original task, we propose to aggregate \textit{vertically}. That is, only the children of an $L_1$ criterion are used to train the regressor.
Additionally, this can have influence on the practical application. In a horizontal aggregation, the user will need to evaluate \textit{all} sub-criteria, even if he intends to only evaluate one of the $L_1$ criteria.

In Table \ref{tab:verticalHorizontal} we showcase the difference between using vertical and horizontal aggregation. Overall, horizontal aggregation outperforms vertical aggregation. This can again be explained by the performance gains coming from the aggregator. When using all sub-criteria as inputs, the regressor has more information , i.e., the classifier has more features, to make a prediction.
However, since a vertical aggregation is more in line with the idea of decomposition, all of our results in Table \ref{tab:main_table} are from vertical aggregation. The HD-Eval result taken from the publication are horizontally aggregated.

\subsection{Floats, Ties, and Rounding} \label{sec:appendix_floatsTiesRounding}

\begin{table*}[tbh]
\centering
\resizebox{0.95\textwidth}{!}{%
\begin{tabular}{lllll|ccc|ccc}
\hline
 & Human &  & With & Outputs & \multicolumn{3}{c|}{SummEval} & \multicolumn{3}{c}{TopicalChat} \\
Method & labels & ICL & rounding & floats & $\rho$ & AP & WR & $\rho$ & AP & WR \\ \hline
\multirow{2}{*}{ICL Decomp} & yes & no & yes & no & 0.567 & 0.842 & \textbf{1.0} & 0.575 & {\ul 0.840} & \textbf{1.0} \\
& yes & no & no & yes & 0.569 & 0.357 & 0.5 & 0.594 & 0.443 & {\ul 0.75} \\
\multirow{2}{*}{AOI Decomp} & yes & no & yes & no & \textbf{0.586} & 0.831 & \textbf{1.0} & 0.574 & 0.839 & 0.33 \\
& yes & no & no & yes & {\ul 0.575} & 0.349 & 0.5 & 0.612 & 0.426 & 0 \\ \hline
\multirow{2}{*}{Direct Prediction w/o float} & yes & yes & yes & no & {0.553} & {0.825} & {\textbf{1.0}} & {{\ul 0.738}} & {\textbf{0.870}} & {0.5} \\
& yes & yes & no & yes & {0.552} & {0.341} & {0.5} & {{\ul 0.738}} & {0.440} & {0.25} \\
\multirow{2}{*}{Direct Prediction w/ float} & yes & yes & yes & no & 0.545 & 0.845 & {\ul 0.75} & 0.716 & \textbf{0.870} & 0.58 \\
& yes & yes & no & yes & 0.565 & 0.331 & 0.5 & \textbf{0.754} & 0.444 & 0.5 \\ \hline
\end{tabular}
} 
\caption{Comparison of performance with and without rounding the outputs.}
\label{tab:results_rounding}
\end{table*}

When calculating alignment between LLM and humans, the de-facto standard is to use the average of three annotators as ground-truth, which produces floating-point values whenever annotators disagree. Individual annotators and standard direct predictions only output integers, but any method using a regressor or aggregator (e.g., HD-Eval) will produce floating-point outputs.

This has two non-obvious consequences. First, a floating-point model can theoretically achieve perfect RMSE against a floating-point ground-truth, while integer-outputting humans cannot. A similar minor benefit arises for Pearson's $r$. Second, floating-point outputs virtually eliminate ties, which affects both the alt-test (where ties count as wins) and Spearman's $\rho$ (which assigns average ranks to tied values). Given the small rating scales (1--5 and 1--3), integer predictions inevitably produce many ties; floating-point predictions do not.

To control for this, we report results with and without rounding (Table~\ref{tab:results_rounding}). Rounding generally improves alt-test AP and WR (by restoring expected tie behavior) but slightly worsens correlations (by removing the numerical advantage of float-to-float comparison). We apply rounding throughout this work to ensure methods are compared fairly: without it, small numerical artifacts (e.g., a score of 4.0002 not tying with ground-truth 4.0) can lead to inflated conclusions. This reveals another reason why the aggregation step in HD-Eval contributes to its reported improvements: it enables floating-point outputs, which benefit from these numerical effects rather than from genuinely better judgment.

\subsection{Regressor Implementation Details}
\label{sec:aggregator-implementation-details}

We implemented the aggregators with scikit-learn using the default parameters:

  \begin{sloppypar}
  \begin{itemize}
      \item \texttt{LinearRegression(*, fit\_intercept=True)}

      \item \texttt{DecisionTreeRegressor(*, criterion='squared\_error', splitter='best',\allowbreak
  max\_depth=None, min\_samples\_split=2, min\_samples\_leaf=1)}

      \item \texttt{RandomForestRegressor\allowbreak(n\_estimators=100, \allowbreak *, criterion='squared\_error',
  max\_depth=None, min\_samples\_split=2, min\_samples\_leaf=1)}

      \item \texttt{MLPRegressor(loss='squared\_error', hidden\_layer\_sizes=(100,),
  activation='relu', *, solver='adam', alpha=0.0001)}
  \end{itemize}
    \end{sloppypar}

\section{Results with other Models} \label{sec:appendix_otherModels}
Table \ref{tab:results_otherModels} shows results using Qwen3-32B \cite{yang2025qwen3} and GPT-OSS-120B~\cite{agarwal2025gpt}. 
Compared to the results from Claude-4 we see that the observation of decomposition not leading to significant performance improvements still hold. This is clear on the TopicalChat dataset. On SummEval, the ICL and AOI decomposition that were proposed as part of this work achieve the best scores. On the AP metric, the benefit from them is smaller than on the correlation metrics.
In general, these two models perform comparatively to Claude-4. There is no clear trend that larger models automatically lead to better performance.

\begin{table*}[tbh]
\centering
\resizebox{0.95\textwidth}{!}{%
\begin{tabular}{lllll|clcc|clcc}
\hline
 &  &  &  & Outputs & \multicolumn{4}{c|}{SummEval} & \multicolumn{4}{c}{TopicalChat} \\
Method & Model & Human Labels & ICL & floats & $\rho$ & $\tau$ & AP & WR & $\rho$ & $\tau$ & AP & WR \\ \hline
\multirow{2}{*}{HD-Eval} & Qwen & yes & no & yes & 0.543 & 0.488 & 0.832 & \textbf{1.0} & 0.570 & 0.500 & {\ul 0.831} & 0.5 \\
 & GPT-OSS & yes & no & yes & 0.560 & 0.505 & 0.847 & \textbf{1.0} & 0.530 & 0.465 & 0.804 & 0.25 \\
\multirow{2}{*}{CheckEval} & Qwen & no & no & yes & 0.437 & 0.368 & - & - & 0.382 & 0.317 & - & - \\
 & GPT-OSS & no & no & yes & 0.411 & 0.352 & - & - & 0.373 & 0.314 & - & - \\ \hline
\multirow{2}{*}{ICL Decomposition} & Qwen & yes & no & yes & 0.563 & 0.510 & \textbf{0.856} & \textbf{1.0} & 0.575 & 0.499 & 0.798 & 0.83 \\
 & GPT-OSS & yes & no & yes & {{\ul 0.567}} & {\ul 0.512} & {0.835} & {\textbf{1.0}} & {0.495} & 0.443 & {0.810} & {\textbf{1}} \\
\multirow{2}{*}{AOI Decomposition} & Qwen & yes & no & yes & {\textbf{0.580}} & \textbf{0.522} & {{\ul 0.850}} & {\textbf{1.0}} & {0.488} & 0.431 & {0.801} & {0} \\
 & GPT-OSS & yes & no & yes & 0.565 & 0.510 & {\ul 0.850} & \textbf{1.0} & 0.507 & 0.445 & 0.801 & 0.16 \\ \hline
\multirow{4}{*}{Direct Prediction} & Qwen & no & yes & yes & 0.525 & 0.461 & 0.629 & 0.5 & 0.558 & 0.492 & 0.813 & 0.33 \\
 & Qwen & yes & yes & yes & 0.541 & 0.488 & 0.844 & \textbf{1.0} & 0.577 & 0.510 & {\ul 0.831} & 0.58 \\
 & GPT-OSS & no & yes & yes & 0.523 & 0.450 & 0.583 & 0.5 & \textbf{0.694} & \textbf{0.614} & 0.790 & 0.33 \\
 & GPT-OSS & yes & yes & yes & 0.496 & 0.448 & 0.828 & \textbf{1.0} & {\ul 0.655} & {\ul 0.587} & \textbf{0.867} & \textbf{1} \\ \hline
\end{tabular}
} 
\caption{Results with Qwen3-32B and GPT-OSS-120B. We show the results averaged across all criteria. We report sample-wise Spearman's $\rho$, Kendall's $\tau$, Advantage Probability (AP), and Win-Rate (WR). Column "ICL" refers to ICL in the evaluation-phase, not the decomposition-phase.}
\label{tab:results_otherModels}
\end{table*}

\section{Per-criterion results} \label{sec:appendix_perCritResult}
We show the per-criterion results in Tables~\ref{tab:result_perdim_summeval} and~\ref{tab:result_perdim_topical}.

\begin{sidewaystable*}
\centering
\resizebox{0.95\textwidth}{!}{%
\begin{tabular}{lllllccccc|ccccc|ccccc|ccccc|ccccc}
\hline
 & Aggre- &  &  & Outputs & \multicolumn{5}{c|}{Coherence} & \multicolumn{5}{c|}{Consistency} & \multicolumn{5}{l|}{Fluency} & \multicolumn{5}{l|}{Relevance} & Average &  &  &  &  \\
Method & gator & Human Labels & ICL & Floats & $r$ & $\rho$ & $\tau$ & AP & WR & $r$ & $\rho$ & $\tau$ & AP & WR & $r$ & $\rho$ & $\tau$ & AP & WR & $r$ & $\rho$ & $\tau$ & AP & WR & $r$ & $\rho$ & $\tau$ & AP & WR \\ \hline
HD-Eval $^\ast$ & MLP & yes & no & yes & \textbf{0.668} & {\ul 0.657} & - & - & - & 0.604 & 0.451 & - & - & - & 0.580 & 0.435 & - & - & - & \textbf{0.619} & \textbf{0.599} & - &  &  & 0.617 & 0.535 & - & - & - \\
HD-Eval (Claude 4) & MLP & yes & no & yes & {0.632} & {0.629} & {\ul 0.533} & {\textbf{0.781}} & {\textbf{1.0}} & {0.715} & {\textbf{0.690}} & {\textbf{0.657}} & {{\ul 0.898}} & {\textbf{1.0}} & {\ul 0.626} & 0.550 & 0.521 & 0.823 & \textbf{1.0} & 0.429 & 0.397 & 0.342 & 0.830 & \textbf{1.0} & 0.601 & 0.567 & {\ul 0.513} & 0.833 & \textbf{1.0} \\
CheckEval $^\ast $ (Mistral-Large) & Mean & no & no & yes & - & {0.644} & \textbf{0.542} & {-} & {-} & {-} & {0.613} & {0.567} & {-} & {-} & - & 0.456 & 0.393 & - & - & - & 0.481 & {\ul 0.417} & - & - & - & 0.549 & 0.480 & - & - \\
CheckEval$^\ast $ (GPT-4o) & Mean & no & no & yes & {-} & {0.556} & {0.464} & {-} & {-} & {-} & {0.530} & {0.474} & {-} & {-} & - & 0.470 & 0.413 & - & - & - & 0.460 & 0.400 & - & - & - & 0.504 & 0.438 & - & - \\
CheckEval (Claude 4) & Mean & no & no & yes & 0.277 & 0.248 & 0.195 & {-} & {-} & 0.511 & 0.465 & 0.400 & {-} & {-} & 0.397 & 0.413 & 0.346 & - & - & 0.524 & 0.520 & \textbf{0.418} & - & - & 0.427 & 0.411 & 0.340 & - & - \\ \hline
ICL Decomp & MLP & yes & no & yes & 0.578 & 0.575 & 0.491 & 0.764 & \textbf{1.0} & 0.703 & 0.664 & 0.631 & 0.889 & \textbf{1.0} & 0.571 & {\ul 0.580} & \textbf{0.543} & \textbf{0.880} & \textbf{1.0} & 0.460 & 0.451 & 0.388 & {\ul 0.836} & \textbf{1.0} & 0.578 & 0.567 & {\ul 0.513} & {\ul 0.842} & \textbf{1.0} \\
AOI Decomp & LR & yes & no & yes & 0.616 & 0.613 & 0.519 & {\ul 0.777} & \textbf{1.0} & 0.715 & 0.667 & 0.632 & 0.883 & \textbf{1.0} & 0.529 & \textbf{0.583} & {\ul 0.541} & 0.844 & \textbf{1.0} & 0.501 & 0.481 & 0.404 & 0.82 & \textbf{1.0} & 0.590 & \textbf{0.586} & \textbf{0.524} & 0.831 & \textbf{1.0} \\
ICL Decomp $ \ddagger$ & LR & yes & no & no & 0.642 & 0.628 & 0.482 & 0.57 & \textbf{1.0} & 0.690 & 0.581 & 0.527 & 0.093 & 0.0 & 0.596 & 0.532 & 0.430 & 0.172 & 0.0 & 0.542 & {\ul 0.536} & 0.399 & 0.592 & \textbf{1.0} & {\ul 0.618} & {\ul 0.569} & 0.459 & 0.357 & 0.5 \\
AOI Decomp $ \ddagger$ & LR & yes & no & no & {\ul 0.662} & \textbf{0.664} & 0.507 & 0.557 & \textbf{1.0} & \textbf{0.726} & 0.606 & 0.552 & 0.072 & 0.0 & 0.619 & 0.498 & 0.403 & 0.175 & 0.0 & {\ul 0.555} & 0.533 & 0.395 & 0.591 & \textbf{1.0} & \textbf{0.640} & 0.575 & 0.465 & 0.349 & 0.5 \\ \hline
Baseline w/o ICL & no & no & no & no & 0.633 & 0.632 & 0.514 & 0.742 & \textbf{1.0} & 0.670 & 0.660 & 0.624 & 0.858 & \textbf{1.0} & 0.375 & 0.423 & 0.378 & 0.15 & 0.0 & 0.457 & 0.444 & 0.377 & 0.477 & 0.0 & 0.534 & 0.540 & 0.473 & 0.557 & 0.5 \\
Baseline w/o ICL & DT & yes & no & yes & 0.586 & 0.576 & 0.494 & 0.763 & \textbf{1.0} & 0.682 & 0.648 & 0.621 & \textbf{0.902} & \textbf{1.0} & 0.472 & 0.453 & 0.429 & 0.735 & 0.0 & 0.462 & 0.432 & 0.373 & 0.789 & \textbf{1.0} & 0.550 & 0.527 & 0.479 & 0.797 & 0.75 \\
Baseline & no & no & yes & no & 0.628 & 0.617 & 0.494 & 0.735 & \textbf{1.0} & 0.699 & 0.669 & 0.630 & 0.866 & \textbf{1.0} & 0.417 & 0.431 & 0.381 & 0.212 & 0.0 & 0.501 & 0.485 & 0.405 & 0.604 & 0.0 & 0.561 & 0.551 & 0.477 & 0.600 & 0.5 \\
Baseline & DT & yes & yes & yes & 0.609 & 0.603 & 0.516 & 0.772 & \textbf{1.0} & 0.700 & 0.666 & 0.636 & \textbf{0.902} & \textbf{1.0} & 0.556 & 0.466 & 0.441 & 0.792 & \textbf{1.0} & 0.509 & 0.477 & 0.411 & 0.835 & \textbf{1.0} & 0.594 & 0.553 & 0.501 & 0.825 & \textbf{1.0} \\
Float Prediction & no & no & yes & yes & 0.623 & 0.611 & 0.488 & 0.759 & \textbf{1.0} & 0.712 & 0.674 & 0.635 & 0.874 & \textbf{1.0} & 0.475 & 0.445 & 0.397 & 0.233 & 0.0 & 0.502 & 0.479 & 0.401 & 0.684 & \textbf{1.0} & 0.578 & 0.552 & 0.480 & 0.638 & 0.75 \\
Float Prediction & DT & yes & yes & yes & 0.609 & 0.599 & 0.512 & 0.775 & \textbf{1.0} & 0.690 & {\ul 0.683} & {\ul 0.648} & 0.892 & \textbf{1.0} & \textbf{0.633} & 0.483 & 0.458 & {\ul 0.874} & \textbf{1.0} & 0.466 & 0.411 & 0.355 & \textbf{0.842} & \textbf{1.0} & 0.599 & 0.544 & 0.493 & \textbf{0.846} & \textbf{1.0} \\
Float Prediction $\ddagger$ & LR & yes & yes & yes & 0.653 & 0.636 & 0.491 & 0.547 & \textbf{1.0} & {\ul 0.718} & 0.646 & 0.601 & {0.052} & 0.0 & 0.493 & 0.466 & 0.393 & 0.168 & 0.0 & 0.537 & 0.511 & 0.401 & 0.583 & \textbf{1.0} & 0.601 & 0.565 & 0.472 & 0.338 & 0.5 \\ \hline
\end{tabular}
} 
\caption{Per-criterion results on SummEval. Highest per column is bolded, second-highest underlined.  $\ddagger :$ Without rounding before comparison.}
\label{tab:result_perdim_summeval}
\end{sidewaystable*}

\begin{sidewaystable*}[]
\centering
\resizebox{0.95\textwidth}{!}{%
\begin{tabular}{lllllccccc|ccccc|ccccc|ccccc|ccccc}
\hline
 & Aggre- &  &  & Outputs & \multicolumn{5}{c|}{Naturalness} & \multicolumn{5}{c|}{Coherence} & \multicolumn{5}{l|}{Engagingness} & \multicolumn{5}{l|}{Groundedness} & \multicolumn{5}{c}{Average} \\
Method & gator & Human Labels & ICL & Floats & $r$ & $\rho$ & $\tau$ & AP & WR & $r$ & $\rho$ & $\tau$ & AP & WR & $r$ & $\rho$ & $\tau$ & AP & WR & $r$ & $\rho$ & $\tau$ & AP & WR & $r$ & $\rho$ & $\tau$ & AP & WR \\ \hline
HD-Eval $^\ast$ & MLP & yes & no & yes & 0.648 & {\ul 0.674} & - & - & - & 0.584 & 0.607 & - & - & - & 0.682 & 0.701 & - & - & - & 0.549 & 0.568 & - & - & - & 0.616 & 0.638 & - & - & - \\
HD-Eval (Claude 4) & RF & yes & no & yes & {0.537} & {0.513} & {0.447} & {0.843} & {\textbf{1.0}} & {0.587} & {0.590} & {0.515} & {0.852} & {0.33} & 0.703 & 0.712 & \textbf{0.712} & \textbf{0.885} & \textbf{1.0} & 0.429 & 0.435 & 0.407 & 0.811 & 0.0 & 0.564 & 0.563 & 0.498 & 0.848 & 0.58 \\
CheckEval $^\ast $ (Mistral-Large) & Mean & no & no & yes & 0.666 & {0.651} & {-} & {-} & {-} & 0.617 & {0.627} & {-} & {-} & {-} & 0.721 & 0.722 & - & - & - & 0.577 & 0.581 & - & - & - & 0.645 & 0.645 & - & - & - \\
CheckEval$^\ast$ (GPT-4o) & Mean & no & no & yes & {0.645} & {0.646} & {-} & {-} & {-} & {0.580} & {0.589} & {-} & {-} & {-} & 0.735 & 0.736 & - & - & - & 0.576 & 0.587 & - & - & - & 0.634 & 0.634 & - & - & - \\
CheckEval (Claude 4) & Mean & no & no & yes & 0.491 & 0.485 & 0.383 & {-} & {-} & 0.624 & 0.627 & 0.534 & {-} & {-} & 0.306 & 0.334 & 0.256 & - & - & 0.259 & 0.264 & 0.220 & - & - & 0.420 & 0.428 & 0.348 & - & - \\ \hline
ICL Decomp & RF & yes & no & yes & 0.499 & 0.506 & 0.432 & 0.824 & \textbf{1.0} & 0.649 & 0.633 & 0.547 & 0.852 & \textbf{1.0} & 0.577 & 0.578 & 0.497 & 0.824 & \textbf{1.0} & 0.578 & 0.581 & 0.543 & 0.861 & \textbf{1.0} & 0.543 & 0.575 & 0.505 & 0.840 & \textbf{1.0} \\
AOI Decomp & LR & yes & no & yes & 0.433 & 0.388 & 0.330 & 0.776 & 0.0 & 0.679 & 0.684 & 0.590 & {\ul 0.861} & {\ul 0.67} & 0.604 & 0.625 & 0.539 & 0.841 & {\ul 0.67} & 0.628 & 0.600 & 0.561 & 0.878 & 0.0 & 0.586 & 0.574 & 0.505 & 0.839 & 0.33 \\
ICL Decomp $ \ddagger$ & RF & yes & no & no & 0.546 & 0.500 & 0.369 & 0.52 & \textbf{1.0} & 0.663 & 0.643 & 0.512 & 0.509 & \textbf{1.0} & 0.623 & 0.651 & 0.494 & 0.511 & \textbf{1.0} & 0.589 & 0.581 & 0.495 & 0.233 & 0 & 0.605 & 0.594 & 0.467 & 0.443 & {\ul 0.75} \\
AOI Decomp $ \ddagger$ & LR & yes & no & no & 0.495 & 0.495 & 0.495 & 0.511 & 0.0 & \textbf{0.737} & \textbf{0.739} & 0.593 & 0.491 & 0.0 & 0.632 & 0.662 & 0.539 & 0.467 & 0.0 & 0.637 & 0.637 & 0.518 & 0.233 & 0.0 & 0.625 & 0.612 & 0.496 & 0.425 & 0.0 \\ \hline
Baseline w/o ICL & no & no & no & no & {\ul 0.664} & \textbf{0.686} & \textbf{0.580} & 0.769 & 0.0 & 0.688 & 0.687 & 0.589 & 0.856 & 0.33 & 0.712 & 0.723 & 0.627 & 0.839 & 0.33 & 0.749 & 0.749 & 0.700 & 0.944 & \textbf{1.0} & 0.703 & 0.711 & 0.624 & 0.852 & 0.42 \\
Baseline w/o ICL & DT & yes & no & yes & 0.547 & 0.559 & 0.493 & 0.839 & \textbf{1.0} & 0.632 & 0.642 & 0.565 & 0.843 & \textbf{1.0} & 0.583 & 0.603 & 0.528 & 0.841 & \textbf{1.0} & 0.749 & 0.749 & 0.700 & 0.944 & \textbf{1.0} & 0.628 & 0.638 & 0.572 & 0.867 & \textbf{1.0} \\
Baseline & no & no & yes & no & 0.652 & 0.663 & {\ul 0.567} & 0.793 & 0.0 & 0.661 & 0.669 & 0.574 & 0.848 & 0.33 & \textbf{0.770} & \textbf{0.785} & {\ul 0.688} & 0.861 & {\ul 0.67} & \textbf{0.862} & \textbf{0.837} & \textbf{0.783} & \textbf{0.978} & \textbf{1.0} & {\ul 0.736} & {\ul 0.738} & \textbf{0.653} & 0.870 & 0.5 \\
Baseline & LR & yes & yes & yes & 0.599 & 0.609 & 0.537 & \textbf{0.863} & \textbf{1.0} & 0.607 & 0.628 & 0.553 & 0.833 & \textbf{1.0} & 0.617 & 0.644 & 0.564 & 0.859 & \textbf{1.0} & \textbf{0.862} & \textbf{0.837} & \textbf{0.783} & \textbf{0.978} & \textbf{1.0} & 0.671 & 0.680 & 0.609 & {\ul 0.883} & \textbf{1.0} \\
Float Prediction & no & no & yes & yes & 0.634 & 0.651 & 0.554 & 0.794 & 0.0 & 0.729 & 0.722 & \textbf{0.634} & \textbf{0.887} & \textbf{1.0} & 0.674 & 0.684 & 0.598 & 0.824 & 0.33 & 0.813 & 0.809 & {\ul 0.756} & {\ul 0.972} & \textbf{1.0} & 0.712 & 0.716 & 0.636 & 0.870 & 0.58 \\
Float Prediction & LR & yes & yes & yes & 0.605 & 0.613 & 0.540 & {\ul 0.852} & \textbf{1.0} & 0.657 & 0.665 & 0.587 & 0.856 & \textbf{1.0} & 0.644 & 0.662 & 0.581 & {\ul 0.872} & \textbf{1.0} & 0.813 & 0.809 & {\ul 0.756} & {\ul \textbf{0.972}} & \textbf{1.0} & 0.680 & 0.687 & 0.616 & \textbf{0.888} & \textbf{1.0} \\
Float Prediction $\ddagger$ & DT & yes & yes & yes & \textbf{0.676} & 0.671 & 0.540 & 0.535 & \textbf{1.0} & {\ul 0.731} & {\ul 0.737} & {\ul 0.614} & {0.472} & 0.33 & {\ul 0.766} & {\ul 0.778} & 0.656 & 0.533 & {\ul 0.67} & {\ul 0.860} & {\ul 0.831} & 0.750 & 0.233 & \textbf{0.0} & \textbf{0.758} & \textbf{0.754} & {\ul 0.640} & 0.443 & 0.5 \\ \hline
\end{tabular}
} 
\caption{Per-criterion results on TopicalChat. Highest per column is bolded, second-highest underlined.  $\ddagger :$ Without rounding before comparison.}
\label{tab:result_perdim_topical}
\end{sidewaystable*}

\section{Prompts}
\begin{figure}[t]
    \centering
    \includegraphics[width=1\linewidth]{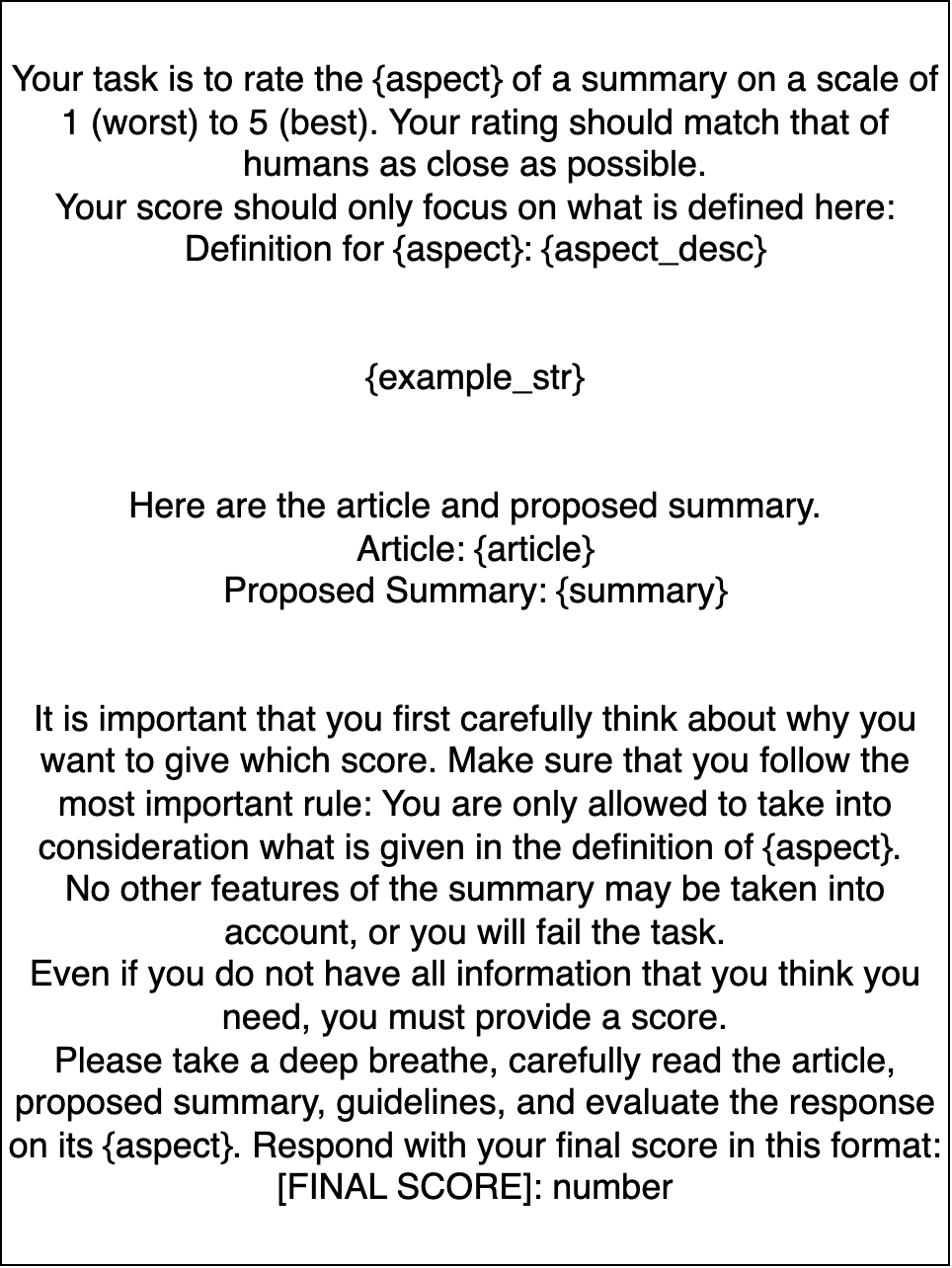}
    \caption{Example Prompt for direct prediction.}
    \label{fig:prompt_direct}
\end{figure}

\begin{figure}[t]
    \centering
    \includegraphics[width=1\linewidth]{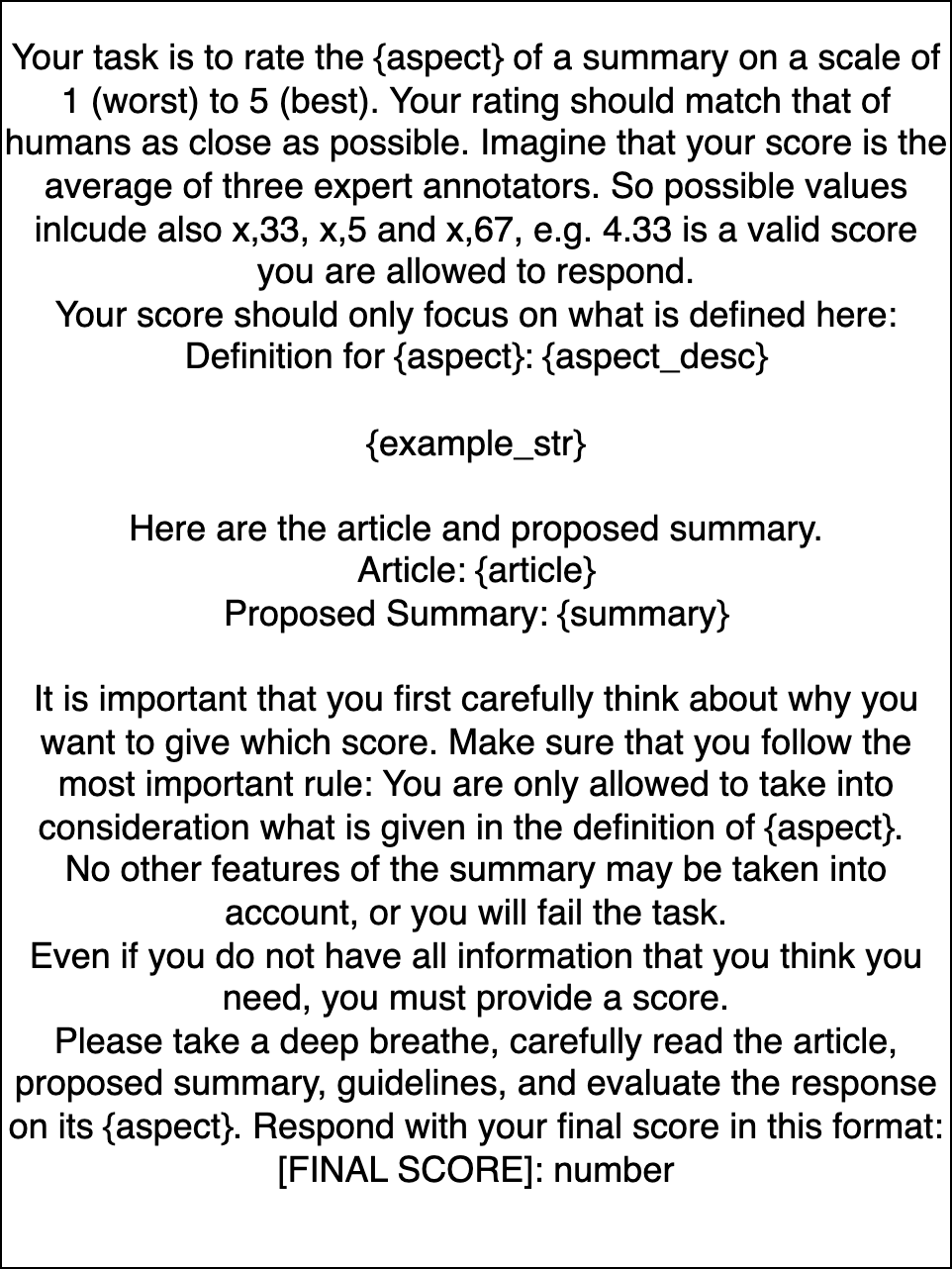}
    \caption{Example Prompt for direct prediction, predicting floating points.}
    \label{fig:prompt_float}
\end{figure}

\begin{figure}[t]
    \centering
    \includegraphics[width=1\linewidth]{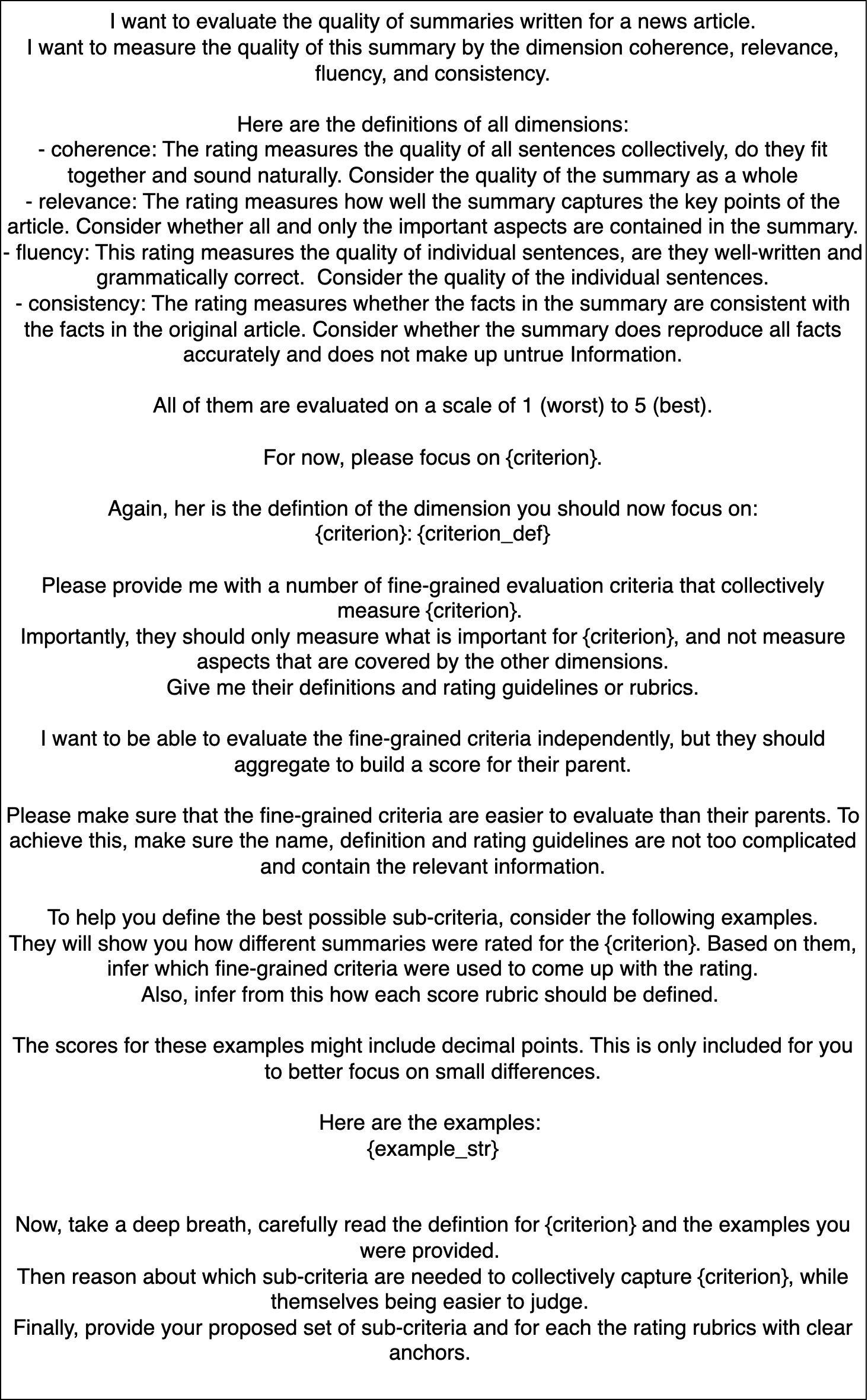}
    \caption{Example Prompt for performing decomposition.}
    \label{fig:prompt_doDecomposition}
\end{figure}

\begin{figure}[t]
    \centering
    \includegraphics[width=1\linewidth]{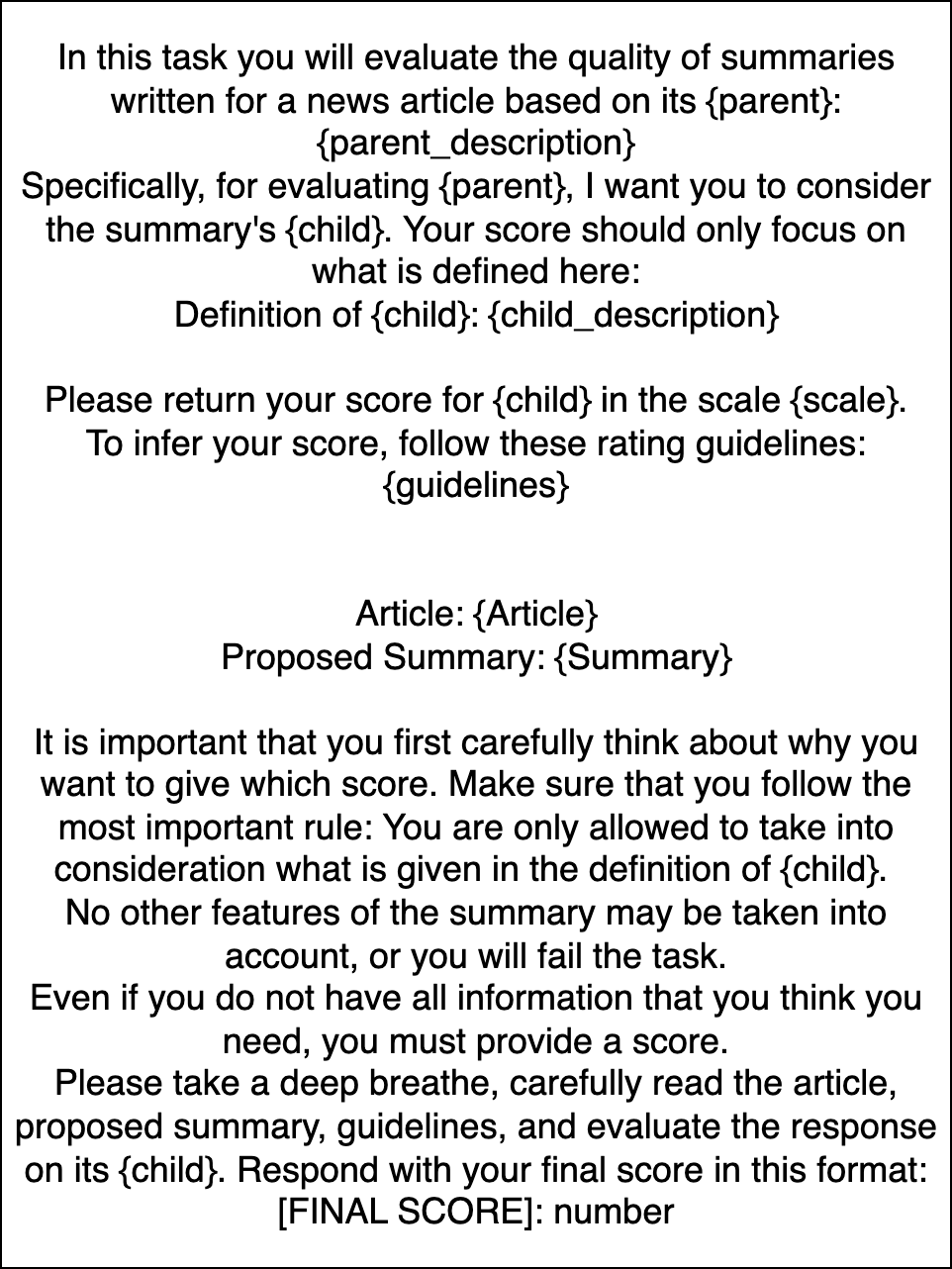}
    \caption{Example Prompt for rating a sub-criteria obtained from decomposition.}
    \label{fig:prompt_decompRate}
\end{figure}

\begin{figure}[t]
    \centering
    \includegraphics[width=1\linewidth]{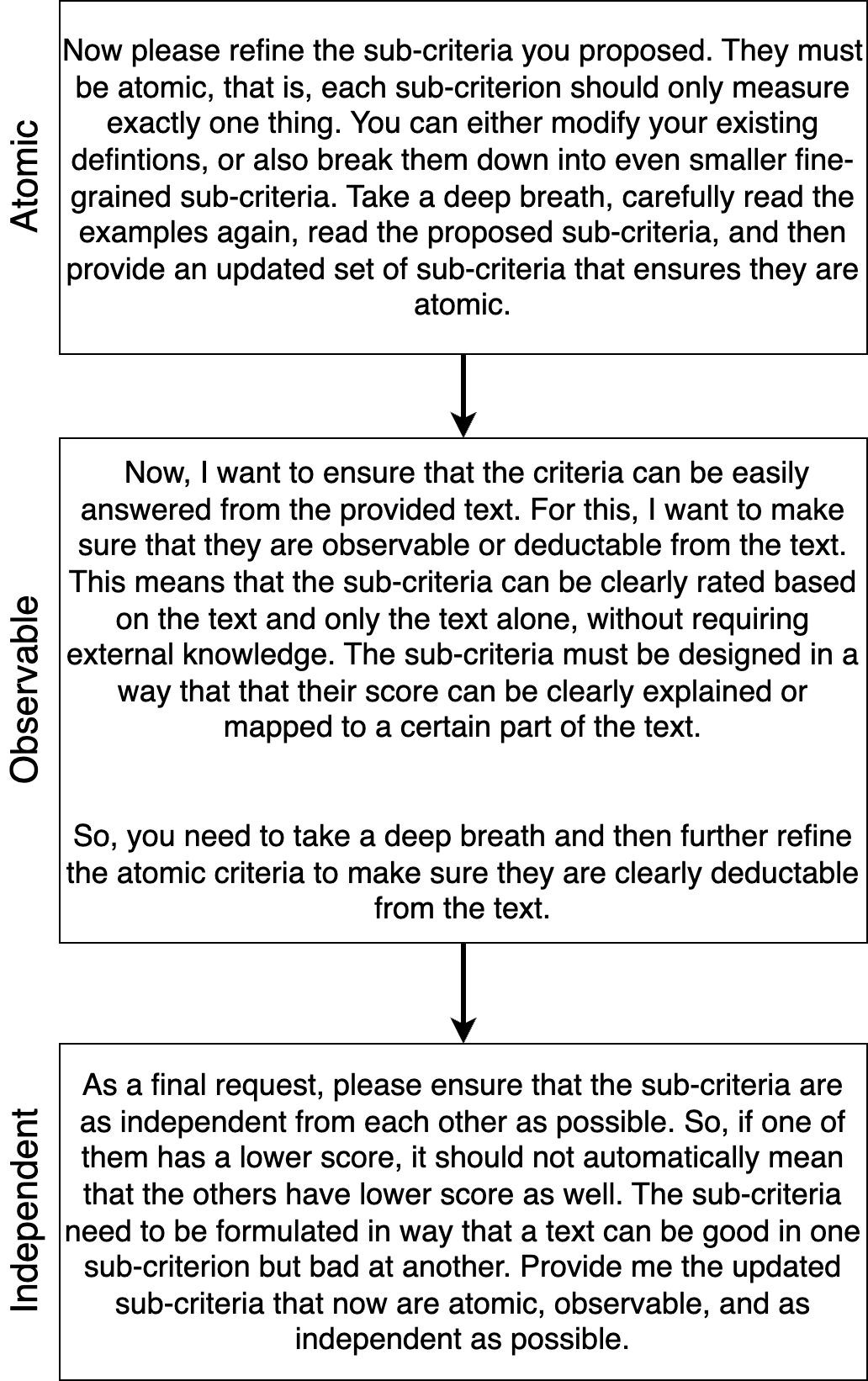}
    \caption{Example Prompt for creating AOI decomposition}
    \label{fig:prompt_decompAOI}
\end{figure}

We show the full prompts in Figures \ref{fig:prompt_direct}, \ref{fig:prompt_float}, \ref{fig:prompt_doDecomposition}, \ref{fig:prompt_decompRate}, and \ref{fig:prompt_decompAOI}.   

\section{AI Assistant Usage}
We used AI assistants as coding assistance and for basic proof-reading of our manuscript. We are solely responsible for all content.  

\end{document}